\documentclass[runningheads]{llncs}

\usepackage{eccv}

\usepackage{eccvabbrv}

\usepackage{graphicx}
\usepackage{booktabs}
\usepackage{multirow}
\usepackage{tabularx}
\usepackage{lipsum}
\usepackage[accsupp]{axessibility}  

\usepackage[pagebackref,breaklinks,colorlinks,citecolor=eccvblue]{hyperref}
\usepackage{hyperref}

\usepackage{orcidlink}
\usepackage{xcolor}
\usepackage{amssymb}
\usepackage{bbding}

\begin{document}

\title{DeeDSR: Towards Real-World Image Super-Resolution via Degradation-Aware Stable Diffusion} 

\titlerunning{DeeDSR}

\author{Chunyang Bi\textsuperscript{\textdagger}\inst{1}\and
Xin Luo\textsuperscript{\textdagger}\inst{1} \and
Sheng Shen\textsuperscript{*} \inst{1} \and
Mengxi Zhang\inst{1}\and
Huanjing Yue\inst{1}\and
Jingyu Yang\textsuperscript{*} \inst{1}
}


\authorrunning{C.~Bi et al.}

\institute{
School of Electrical and Information Engineering, Tianjin University, Tianjin, China
\email{\{codyshens, yjy\}@tju.edu.cn}}

\maketitle

\newcommand\blfootnote[1]{%
\begingroup
\renewcommand\thefootnote{}\footnote{#1}%
\addtocounter{footnote}{-1}%
\endgroup
}
\blfootnote{\textdagger Equal contribution.}
\blfootnote{*Corresponding author.}

\begin{abstract}
Diffusion models, known for their powerful generative capabilities, play a crucial role in addressing real-world super-resolution challenges. However, these models often focus on improving local textures while neglecting the impacts of global degradation, which can significantly reduce semantic fidelity and lead to inaccurate reconstructions and suboptimal super-resolution performance. To address this issue, we introduce a novel two-stage, degradation-aware framework that enhances the diffusion model's ability to recognize content and degradation in low-resolution images. In the first stage, we employ unsupervised contrastive learning to obtain representations of image degradations. In the second stage, we integrate a degradation-aware module into a simplified ControlNet, enabling flexible adaptation to various degradations based on the learned representations. Furthermore, we decompose the degradation-aware features into global semantics and local details branches, which are then injected into the diffusion denoising module to modulate the target generation. Our method effectively recovers semantically precise and photorealistic details, particularly under significant degradation conditions, demonstrating state-of-the-art performance across various benchmarks. Codes will be released at \href{https://github.com/bichunyang419/DeeDSR}{https://github.com/bichunyang419/DeeDSR}.

\end{abstract}

\begin{figure}[tb]
  \centering

  \includegraphics[height=6.8cm]{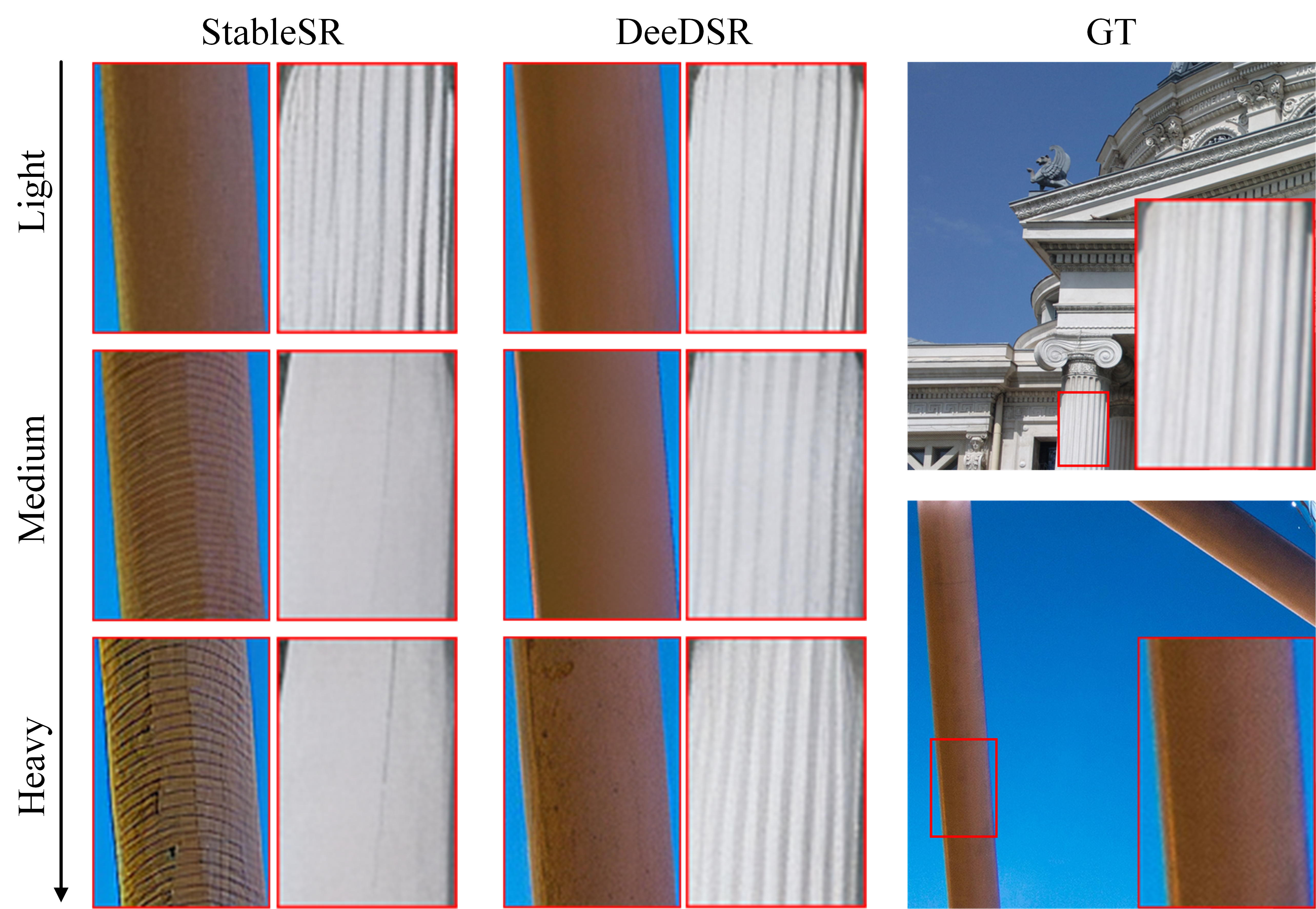}
  \caption{Qualitative comparison of our DeeDSR and StableSR\cite{wang2023stablesr} under various degradation conditions on the synthetic DIV2K dataset. Our model demonstrates robustness against different degradations, generating correct semantics and textures, while StableSR produces incorrect semantics and textures under severe degradation. Degradation levels include light, medium, and heavy, detailed in \cref{sec:ablation} and \cref{fig:degradation}} 
  \vspace{-4mm}
  \label{fig:introduction}
\end{figure}

\section{Introduction}
\label{sec:intro}
In the realm of image processing, super-resolution (SR) stands as a cornerstone task, dedicated to enhancing low-resolution (LR) images to their high-resolution (HR) counterparts. This technology has widespread applications across critical domains such as mobile photography~\cite{chen2019camera,wang2021dualcamera,yang2021implicit}, medical image analysis~\cite{georgescu2023multimodal,li2021medicalreview}, security surveillance~\cite{rasti2016surveillance,farooq2021surveillance} and significantly impacts various computer vision tasks, including object detection~\cite{shermeyer2019detection}, segmentation~\cite{wang2020dual}, and recognition~\cite{gunturk2003eigenface}. Despite recent advancements, SR in complex real-world scenarios remains challenging.  A common strategy involves leveraging image priors, either explicitly through reference images~\cite{jiang2021c2-matching,cao2022reference,lu2021masa} or implicitly via pre-trained generative models~\cite{chan2021glean,menon2020pulse,pan2021exploiting,wang2023stablesr,wang2021towards}. Notably, recent text-to-image (T2I) diffusion models have shown remarkable potential for SR tasks~\cite{wang2023stablesr, lin2023diffbir, yang2023pasd, wu2023seesr}. These diffusion-based SR methods have demonstrated the ability to generate realistic image details, yet they have limitations. StableSR~\cite{wang2023exploring} and DiffBIR~\cite{lin2023diffbir} only rely on LR images for control signals, neglecting the semantic text information in pre-trained T2I models. This leads to inherent limitations in capturing the holistic image context and often fails to restore severely degraded but semantically important details. Moreover, the ill-posed nature of LR images introduces the risk of introducing semantically erroneous textures, as depicted in \cref{fig:introduction}, where increasing degradation levels challenge StableSR's ability to recover correct image semantics. Recognizing these issues, PASD~\cite{yang2023pasd} and SeeSR~\cite{wu2023seesr} attempt to enhance semantic accuracy with text prompts. However, these approaches have two main drawbacks: the text encoder in visual language models~\cite{radford2021clip} is not optimized for low-level tasks, often extracting coarse-grained information that fails to accurately identify degradation information; and the text prompt encoded information does not effectively represent the image degradation process, which includes various degradations such as low-resolution, blurriness, and noise, as shown in \cref{fig:degradation}. The difficulty in precisely describing this process with language limits the method's quality and effectiveness in real-world image reconstruction.

As the adage goes, "a picture is worth a thousand words," a sentiment that resonates across languages and eras, suggesting that images are more expressive than texts from a human perspective. Based on the above analysis, we propose \textbf{DeeDSR}: \textbf{De}gradation-Awar\textbf{e} Stable \textbf{D}iffusion \textbf{SR}, a method that leverages image prompts to generate global degradation representations, thereby enhancing the generative capabilities of pre-trained T2I models in real SR. DeeDSR consists of two stages. The first stage employs contrastive learning to understand degradation semantics; the second stage uses degradation prompts in conjunction with LR images to precisely control the T2I model, fostering the generation of rich and semantically coherent details. Our contributions can be summarized as follows:
\begin{itemize}
\item We introduce DeeDSR, a novel method for real-world SR that leverages degradation-aware image prompts to enhance the generative capabilities of the pre-trained diffusion model, focusing on both global degradation and semantic accuracy.
\item We adopt a two-stage pipeline that first employs contrastive learning to capture degradation representation, followed by the integration of these semantics with LR images to precisely control the T2I model, resulting in the generation of detailed and semantically coherent images.
\item  Extensive experimental validations demonstrate DeeDSR's superiority in recovering high-quality, semantically accurate outputs under diverse degradation conditions, outperforming existing methods.
\end{itemize}

\section{Related work}
\subsection{Real-World Image Super-Resolution}
Research in the field of real-world SR primarily focuses on two main strategies: optimizing data utilization~\cite{cai2019toward,chen2019camera,wei2020component,wang2021realesrgan,zhang2021designing} and integrating image priors~\cite{chan2021glean,menon2020pulse,pan2021exploiting,wang2023stablesr}. The first strategy involves creating diverse and realistic paired data by modifying physical data collection methods~\cite{chan2021glean} or enhancing the data generation pipeline~\cite{wang2021realesrgan,menon2020pulse,pan2021exploiting} and then implicitly or explicitly modeling degradation using this data. The second strategy emphasizes the use of image priors. While methods that start from scratch require a large amount of data and computational resources, it has been proven that using pre-trained generative models~\cite{wang2023stablesr,chan2021glean,gu2020image,menon2020pulse,pan2021exploiting,wang2021towards} with detailed texture priors is a practical and cost-effective approach. Some research utilizes pre-trained GANs~\cite{chan2021glean,menon2020pulse,wang2021towards,gu2020image} to enhance the SR process.
However, due to the inherent limitations of GANs, these methods sometimes generate unrealistic textures. Therefore, in recent research, there is increasing interest in using more advanced pre-trained generative models, such as denoising diffusion models.

\subsection{Diffusion-Based Super-Resolution}
Previously, researchers~\cite{li2022srdiff, saharia2022image} utilized Diffusion Probability Models (DDPM) to tackle image SR, assuming direct downsampling degradation. However, this assumption limited their practicality in complex scenarios like Real SR. These models trained on vast image-text pairs, provide robust image priors crucial for handling Real SR complexities. For instance, StableSR~\cite{wang2023stablesr} refines the SD model via time-aware encoder training and employing feature distortion for balancing fidelity and perceptual quality. Similarly, DiffBIR~\cite{lin2023diffbir} adopts a two-stage strategy: initially using SwinIR to remove degradation (noise, artifacts) for a clean image, followed by a diffusion model for detail replenishment. However, these methods primarily rely on local image information to activate T2I model generative capabilities. In contrast, some works propose utilizing text to represent degradation information~\cite{yang2023pasd, wu2023seesr,yue2024resshift, chen2023textprompt, sun2023coser, yu2024scaling}. For example, PASD~\cite{yang2023pasd} utilizes off-the-shelf high-level semantic models like ResNet~\cite{he2015deep} to extract semantic information guiding diffusion, enhancing T2I model generative power. SeeSR~\cite{wu2023seesr}, extracts text information from LR images as a prior, improving super-resolution model understanding of image content and increasing restoration accuracy.  While these models utilize text to control semantics, they are not totally degradation-aware, especially when complex degradations cannot be effectively described by text. Despite notable visual quality advancements, they have yet to fully harness large text-to-image generation models' potential, mainly due to limited image content comprehension.

\begin{figure}[tb]
  \centering
  \includegraphics[height=6.4cm]{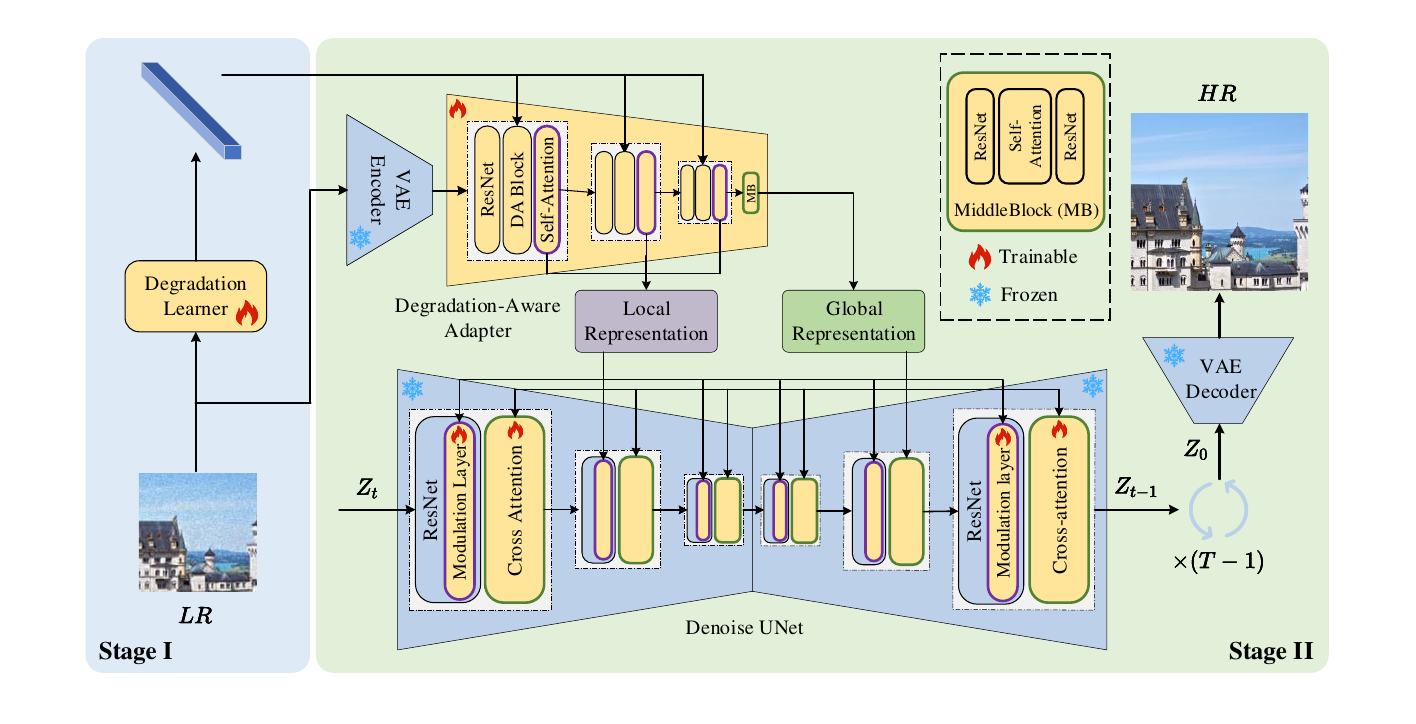}
  \caption{Pipeline of our proposed DeeDSR. In the first stage, we capture global degradation semantics using a contrastive learning strategy. In the second stage, the degradation information is integrated with LR images via degradation-aware (DA) blocks to precisely control the pre-trained Stable Diffusion (SD) model through Cross Attention Modules and Modulation Layers. Note that the Degradation Learner is fixed in the second stage.
  }
  \vspace{-4mm}
  \label{fig:fig2}
\end{figure}

\section{Methodology}
The overall architecture of the proposed method is shown in \cref{fig:fig2}. Our DeeDSR adopts a two-stage pipeline. In the first stage, we utilize a contrastive learning strategy to estimate degradation representations from LR images, as described in \cref{sec:degradation}. In the second stage, we incorporate the degradation representations with LR images to guide the Stable Diffusion (SD) model, as detailed in \cref{sec:degradationinsert}. We start with some preliminaries in the following section.

\subsection{Preliminaries}

In this work, we employ Stable Diffusion~\cite{rombach2022high} as the base diffusion model. First, the input image $x$ is encoded into the latent space by a VAE autoencoder~\cite{rombach2022high} $\mathcal{E}$, \textit{i.e.}, $z=\mathcal{E}(x)$. Then, Gaussian noise is iteratively added, yielding a noisy latent $z_t$ after $t$ steps. The training process aims to estimate the Gaussian noise $\varepsilon$ at step $t$ via a UNet architecture conditioned on texts or images (denoted as $c$). The UNet integrates residual blocks and cross-attention layers at different spatial scales. Text prompts are encoded by a CLIP~\cite{radford2021clip} text encoder, producing multimodal features fed into the UNet through cross-attention layers. Time steps are embedded into the UNet via residual blocks. The optimization can be denoted as:

\begin{equation}
\mathop {\min }\limits_\theta {\mathbb{E}_{{z_t},t,c,\varepsilon }}\left[ {\left\| {\varepsilon  - {\varepsilon _\theta }\left( {z_t},t,c \right)} \right\|_2^2} \right],
\end{equation}
After multiple iterative steps, a clean latent $z_0$ is obtained and subsequently decoded back to the pixel space by a VAE decoder $\mathcal{D}$, \textit{i.e.}, $x' = \mathcal{D}(z_0)$.

To incorporate text information in the image generation process, Stable Diffusion adopts cross-attention layers. Specifically, the latent image feature $\mathcal{F}$ and text feature $\mathcal{F}_c$ encoded by the CLIP text encoder are transformed by projection layers to obtain the query $Q = W_Q \cdot {\mathcal{F}}$, key $K = W_K \cdot {\mathcal{F}_c}$, and value $V = W_V \cdot {\mathcal{F}_c}$, where $W_Q$, $W_K$, and $W_V$ are weight parameters of the query, key, and value projection layers, respectively. Attention is conducted by a weighted sum over value features:
\begin{equation}
\text{Attention}(Q, K, V) = \text{Softmax}\left(\frac{QK^T}{\sqrt{d}}\right)V,
\end{equation}
where $d$ is the output dimension of key and query features. The latent feature is then updated with the output of the attention block.

\subsection{Degradation Learner}
\label{sec:degradation}

Accurately modeling the unknown degradation of low-resolution images is essential for real-world image super-resolution tasks. Chen \etal  \cite{chen2023image} introduced text prompts to provide degradation priors for image super-resolution. Similarly, Qi \etal \cite{qi2023tip} proposed a more detailed degradation description to serve as universal guidance control for image restoration, but it can only handle relatively simple degradation processes. This is because the degradation information provided by text prompts is inherently limited, and complex degradations are difficult to describe in language. Even if described, visual language models struggle to effectively understand and generate corresponding representations. In contrast, images contain more abundant degradation semantic information. Inspired by Wang \etal \cite{wang2021unsupervised}, we adopt a contrastive learning strategy to comprehensively learn degradation representations. Given the complexity of our degradation types, we employ ResNet50 \cite{he2015deep} as the Degradation Learner, a deeper network architecture, to encode the image space into the latent space.


Given that degradation is consistent within each image but varies across different images, we adopt the contrastive learning strategy~\cite{he2020moco_v1, chen2020moco_v2}, which brings features of the same degradation type closer together in the feature space while pushing features of different degradation types apart. We consider an image patch as the query patch, $x^{q}$. The positive samples ($x^{+}$) consist of the patches within the same image, while the negative samples ($x^{-}$) comprise those patches from different images. Then we encoder the query patch, positive and negative samples into degradation representation using ResNet50, $E_{degrad}$. Considering that a queue contains one positive sample, ${x^{+}_0}$, and $N$ negative samples, ${x^{-}_1} ,{x^{-}_2}, \cdots ,{x^{-}_N} $, the encoder process can be denoted as:
\begin{align}
  \mathcal{F}_q & = E_{degrad}(x^q),  \\
\mathcal{F}_{k ^ +}& = E_{degrad}({x^{+}_0}), \\
\mathcal{F}_{{k^{-}_i}} & = E_{degrad}({x^{-}_i}), i=1,2,\cdots,N 
\end{align}
We define the training loss as:
\begin{equation}
  {L_{stageI}} =  - \log \frac{{\exp \left( {\mathcal{F}_q \cdot {\mathcal{F}_{k ^ +} }/\tau } \right)}}{{\sum\limits_{i = 1}^N {\exp \left( {\mathcal{F}_q \cdot \mathcal{F}_{{k^{-}_i}} /\tau } \right)}}}.
  \label{eq:important}
\end{equation}
where $\tau$ is a temperature hyper-parameter.

\subsection{Degradation-Aware Stable Diffusion}
\label{sec:degradationinsert}
Diffusion models are creative but random. ControlNet~\cite{zhang2023adding} adds conditional control by using additional details like depth and edges to guide generation. Therefore, we introduce ControlNet as our conditional control module. Considering that in the SR domain, the model's generated results must possess a certain level of fidelity, directly using image depth, edges, keypoints, and other information cannot adequately adapt to the SR task. We use the LR image as a condition to guide the network's generation. Furthermore, to make the ControlNet perceive degradation, we inject the degradation representation learned in the first stage into ControlNet, allowing ControlNet to better perceive degradation, termed as Degradation-Aware Adapter. Our results shown in~\cref{fig:introduction} demonstrate that introducing the degradation representation into ControlNet enables more robust handling of various degrees of degradation. Simultaneously, to reduce computational complexity, we reduce the number of modules in the original ControlNet, decreasing the number of trainable parameters by approximately one-third. 

Our proposed structure is illustrated in \cref{fig:fig2}. First, the pre-trained latent diffusion UNet is frozen during training, except for the cross-attention layers, to preserve the image prior. Second, our proposed Degradation-Aware Adapter is randomly initialized to learn the restoration conditions. It takes the degraded image and the degradation representation as input and outputs global and local representations. These representations are then embedded into the cross-attention layers and modulation layers of the UNet, respectively, to guide the generation process towards restoring the high-quality image from the degraded input.

\textbf{Global representation.} As discussed above, most diffusion-based methods either keep the text prompt null~\cite{wang2023stablesr,lin2023diffbir} or use text to describe degradation~\cite{wu2023seesr,chen2023textprompt}. Neither of these approaches effectively addresses the issue of degradation representation, which is inherently ill-posed.  Therefore, we propose to replace the text prompts embeddings in cross-attention modules with image prompts embeddings obtained by the Degradation-Aware Adapter. Additionally, the query, key, and value projection layers (\textit{i.e.} $W_Q$, $W_K$, and $W_V$) in the cross-attention layer are turned on during training.



\textbf{Local representation.} Besides, local texture information should also be considered to maintain a balance between realism and fidelity. Therefore,  We incorporate Modulation Layers into the frozen SD UNet, as shown in \cref{fig:fig2}. These layers are designed to modulate intermediate features through multi-scale local representations. The modulation process can be expressed as follows:

\begin{align}
{\gamma ^{{s_i}}},{\beta ^{{s_i}}} &= H_M\left( {{\mathcal{F}^{{s_i}}_L}} \right), i=1,2,\cdots,n \\
\hat {\mathcal{F}}_{res}^{{s_i}} &= \left( {1 + {\gamma ^{{s_i}}}} \right)\mathcal{F}_{res}^{{s_i}} + {\beta ^{{s_i}}} 
\end{align}
where $\gamma ^{{s_i}}, \beta ^{{s_i}}$ are the scaling and bias parameters corresponding to the ${s_i}$ spatial scale. $H_M$ denotes a lightweight network that consists of several convolution layers. $\mathcal{F}_{res}^{{s_i}}$ and $\hat {\mathcal{F}}_{res}^{{s_i}}$ respectively represent the residual features before and after modulation.

\textbf{Training Process.} The training pipeline of our DeeDSR is as follows. We passed the HR images through a pre-trained VAE autoencoder \cite{rombach2022high} to obtain a latent representation, ${z_0}$. Then Gaussian noise was added to ${z_0}$ step by step, finally resulting in a noise latent ${z_t}$ after $t$ steps. With the current time step $t$, multi-scale local representations, ${\mathcal{F}^{s_i}_L}, i=1,2,\cdots,n$ and a global semantic representation, ${\mathcal{F}_G}$, we train our DeeDSR network, denoted as ${\varepsilon _\theta}$, to estimate the noise, $\varepsilon$, added to the noisy latent ${z_t}$. The loss function is:
\begin{equation}
{L_{stageII}} = {\mathbb{E}_{{z_t},t,{z_{lr}},{d_{lr}},\varepsilon }}\left[ {\left\| {\varepsilon  - {\varepsilon _\theta }\left( {{z_t},t,{H_{DA}}\left( {{z_{lr}},{d_{lr}}} \right)} \right)} \right\|_2^2} \right]
\end{equation}
where $z_{lr}$ is the LR latent and $d_{lr}$ is the degradation representation estimated by the pre-trained Degradation Learner. $H_{DA}$ denotes the Degradation-Aware Adapter.

\subsection{Noise Guidance for a tradeoff between realism and fidelity.}
While DeeDSR yields satisfactory SR results, the diverse requirements of various tasks and user preferences necessitate a controllable inference strategy. This strategy allows for the direct incorporation of LR information, facilitating adjustments between realistic and smoothed results. 
A common approach is to directly embed the degraded image (LR) into the initial state \cite{yue2024resshift}. However, this method significantly compromises the quality of SR due to its alignment with the model input format during the training of the SD. This alignment introduces bias in the output SR, favoring features from the LR input.
To address this issue, we propose a more sophisticated guidance method known as Noise Guidance used in the DDIM \cite{song2020DDIM} sampling process due to the role of the $\sqrt{1 - \alpha_{t-1} - \sigma_t^2} \varepsilon _{\theta}^{(t)} (x_t)$ pointing to $x_t$ in its non-Markovian sampling process:
\begin{equation}
x_{t-1} = \sqrt{\alpha_{t-1}} \left(\frac{x_t - \sqrt{1 - \alpha_t} \varepsilon_{\theta}^{(t)}(x_t)}{\sqrt{\alpha_t}} \right) + \sqrt{1 - \alpha_{t-1} - \sigma_t^2} \varepsilon _{\theta}^{(t)} (x_t) + \sigma_t \varepsilon _t 
\end{equation}
where the $\varepsilon _{\theta}^{(t)} (x_t)$ represents the noise predicted by Diffusion Model (DM). Noise guidance involves selecting specific $n$ steps in the sampling process and computing the $\varepsilon _{\theta}^{(t)} (x_t)$ by ${z_{lr \uparrow }}$ (Upsample the LR by a factor of 4 using bicubic interpolation, and then embed it into the latent space by $\mathcal{E}$.). This process can be described as follows:
\begin{equation}
\varepsilon _{\theta}^{(t)} (x_t) = \frac{x_t - \sqrt{\alpha_t} \cdot z_{lr \uparrow } }{\sqrt{1 - \alpha_t}}
\end{equation}
by default, the Noise Guidance is only applied in the first sample from the random noise.

The above guidance maintains the sampling from random noise, which ensures the image generation quality while forcing the spatial structure of the LR to be aligned as well as the color consistency to improve the fidelity. More ablation is provided in the \textbf{supplementary material}.

\section{Experiment}
\begin{figure}[h]
  \centering
  \includegraphics[width=\textwidth]{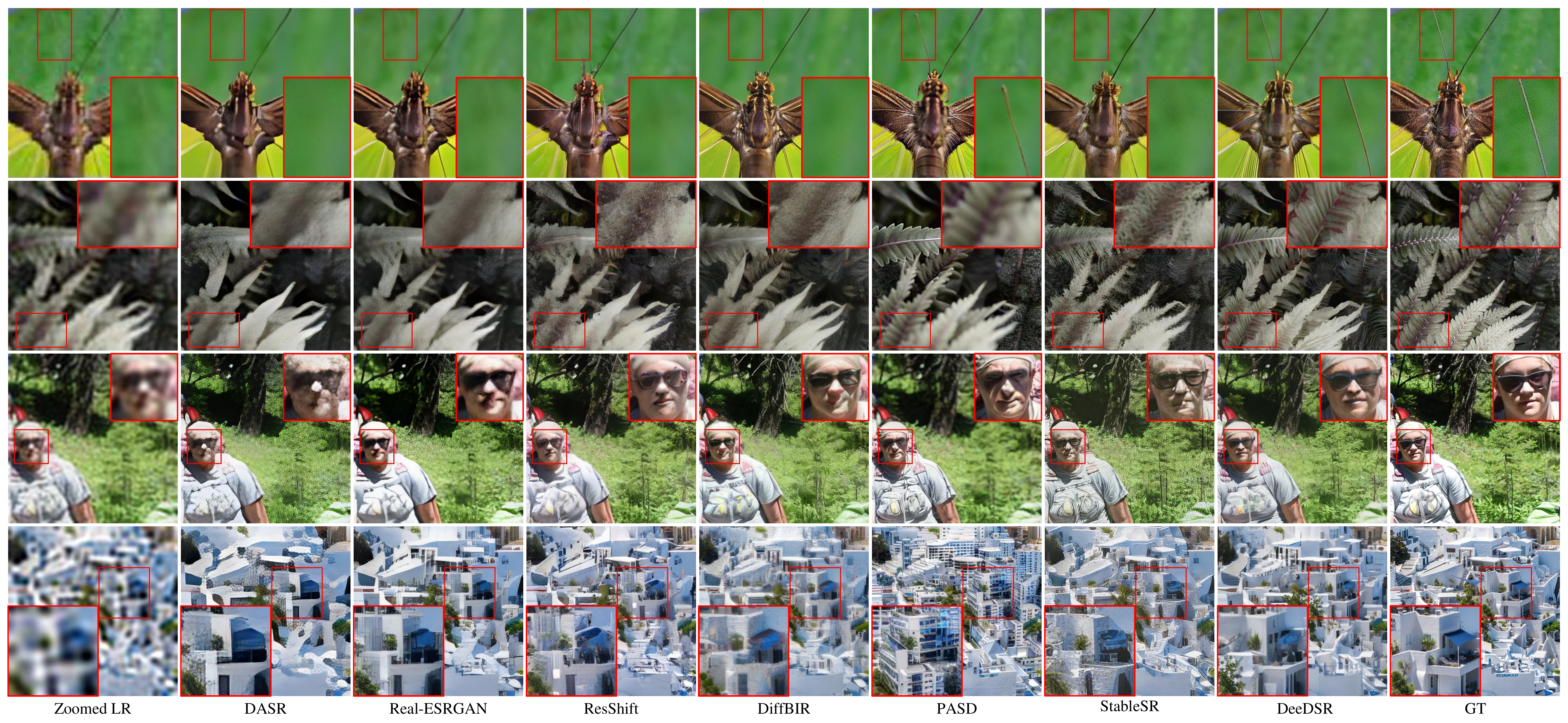}
  \caption{Qualitative comparisons of different Real-ISR methods on synthetic dataset. Please zoom in for a better view.}
  \label{fig:Qualitative Comparisons(synthetic)}
  \vspace{-2mm}
\end{figure}
\begin{figure}[h]
  \centering
  \includegraphics[width=\textwidth]{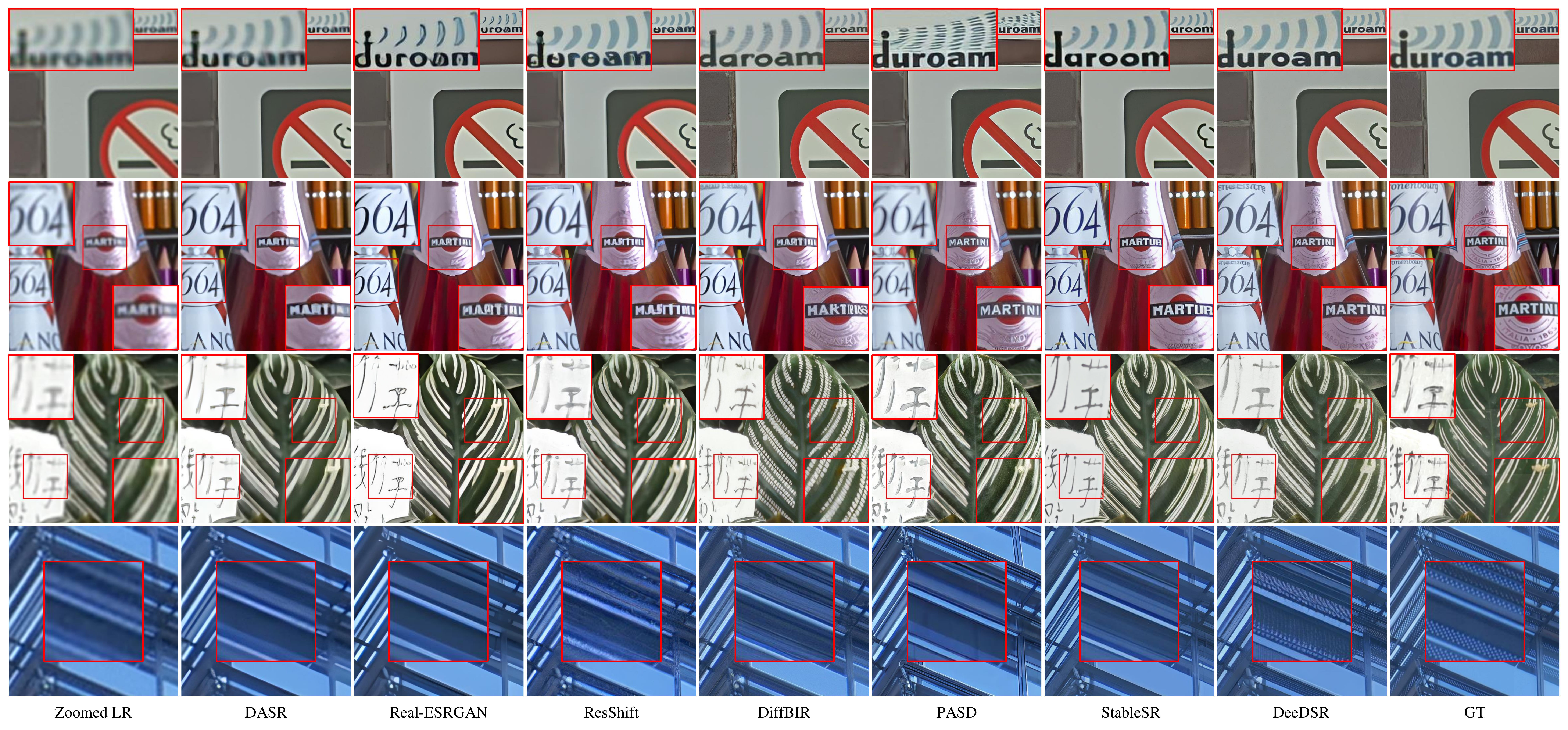}
  \caption{Qualitative comparisons of different methods on real-world datasets. Please zoom in for a better view.}
  \label{fig:Qualitative Comparisons(real)}
\end{figure}
\subsection{Implementation Details}
We adopt a two-stage pipeline. In the first stage, we train the Degradation Learner via contrastive learning with $\tau$ and $N$ in \cref{eq:important} to 0.07 and 16384, respectively. We train on 64$\times$64 LR patches with a batch size of 64, the initial learning rate of 0.03 (decayed by 10 at 120 and 240 epochs), for 630 epochs on  DIV2K \cite{agustsson2017ntire}, DIV8K \cite{gu2019div8k}, Flickr2K\cite{timofte2017ntire}, the first 5k face images from FFHQ\cite{karras2019style}. In the second stage, we use Stable Diffusion 2.1-base \footnote{https://huggingface.co/stabilityai/stable-diffusion-2-1-base} as the pre-trained model, freeze Degradation Learner parameters, and train the randomly initialized Degradation-Aware Adapter from scratch. The whole DeeDSR model is trained on $512\times512$ resolution for 65840 iterations with a batch size of 128, using ADAM \cite{kingma2014adam} optimizer with learning rate $1 \times 10^{-5}$. Our implementation uses PyTorch 1.12.0, CUDA 11.4, on 4 NVIDIA A100 GPUs. For inference, we apply DDIM\cite{song2020denoising} sampling with 50 timesteps.

\subsection{Experimental Settings}
\vspace{-3mm}
\noindent\textbf{Training Datasets.}  To train DeeDSR, we apply DIV2K \cite{agustsson2017ntire}, DIV8K \cite{gu2019div8k}, Flickr2K\cite{timofte2017ntire}, OutdoorSceneTraining\cite{wang2018recovering}, the first 5k face images from FFHQ\cite{karras2019style} and LSDIR\cite{li2023lsdir} as the training dataset. We follow the degradation pipeline in Real-ESRGAN \cite{wang2021realesrgan} to generate the corresponding LR images.

\noindent\textbf{Test Datasets.} We evaluate our model both in synthetic and real-world datasets. For the synthetic dataset, we randomly crop the validation set of DIV2K \cite{agustsson2017ntire} into $512\times512$ resolution and generate 3k LR-HR pairs using the same degradation pipeline as the training dataset. We refer to this dataset as DIV2K-Val. For real-world datasets, we center-crop the LR images from the RealSR\cite{cai2019toward} and DRealSR\cite{wei2020component} datasets to $128\times128$ resolution.

\noindent\textbf{Evaluation Metrics.}To assess the performance of different methods, we utilize a series of metrics, including both reference and non-reference ones. For reference metrics, we adopt PSNR and SSIM \cite{wang2004image} scores (evaluated on the Y channel in YCbCr color space) for fidelity evaluation. LPIPS \cite{zhang2018unreasonable} and FID \cite{heusel2017gans} are employed for perceptual quality measures. For non-reference metrics, we adopt CLIPIQA\cite{wang2023exploring}, MUSIQ\cite{ke2021musiq}, BRISQUE \cite{mittal2012no} and MANIQA \cite{yang2022maniqa} for evalution.

\noindent\textbf{Compared Methods.}We compare DeeDSR with several state-of-the-art methods, including GAN-based methods, BSRGAN\cite{zhang2021designing}, Real-ESRGAN\cite{wang2021realesrgan}, LDL\cite{liang2022details}, DASR\cite{liang2022efficient}, FeMaSR\cite{chen2022real}, and diffusion-based methods, LDM\cite{rombach2022high}, StableSR\cite{wang2023stablesr},ResShift\cite{yue2024resshift}, PASD\cite{yang2023pasd}, DiffBIR\cite{lin2023diffbir}. We use the official codes and released models of these methods to test for fair comparisons.

\begin{table}[t]
\centering
\caption{Quantitative comparison with state-of-the-art methods on both synthetic and real-world benchmarks. The best and second-best
results of each metric are highlighted in red and blue, respectively.}
\renewcommand\arraystretch{1.2}
\resizebox{0.98\textwidth}{!}{%
\begin{tabular}{@{}c|c|ccccc|cccccc@{}}
\hline
\multirow{2}{*}{\textbf{Datasets}} & \multirow{2}{*}{\textbf{Metrics}} & \textbf{BSRGAN} & \textbf{Real-}\cite{wang2021realesrgan} & \textbf{LDL} & \textbf{DASR} & \textbf{FeMaSR} & \textbf{LDM}  & \textbf{StableSR}  & \textbf{ResShift}  & \textbf{PASD} 
& \textbf{DiffBIR} & \multirow{2}{*}{\textbf{DeeDSR}} \\
 &  & \cite{zhang2021designing} & \textbf{ESRGAN} & \cite{liang2022details} & \cite{liang2022efficient} & \cite{chen2022real}& \cite{rombach2022high} & \cite{wang2023stablesr}  & \cite{yue2024resshift} & \cite{yang2023pasd}
& \cite{lin2023diffbir}& 
\\ \hline
\multirow{8}{*}{DIV2K-Val} & PSNR ↑ & \textcolor{blue}{23.00} & 22.91 & 22.41 & 22.96 & 21.49 & 22.06 & 21.85 & \textbf{\textcolor{red}{23.06}} & 21.68 & 22.21 & 22.51 \\ 
 & SSIM ↑ & 0.5802  & \textbf{\textcolor{red}{0.5953}} & \textcolor{blue}{0.5868} & 0.5863 & 0.5408 & 0.4707 & 0.5315 & 0.5739 & 0.5405 & 0.5232 & 0.5538 \\ 
 & LPIPS ↓ & 0.3452  & 0.3177 & 0.3322 & 0.3569 & 0.3314 & 0.4333 & \textbf{\textcolor{red}{0.3132}} & 0.3476 & 0.3686 & 0.3554 & \textcolor{blue}{0.3133} \\ 
 & FID ↓ & 48.52 & 41.63  & 47.51 & 53.84 & 41.42 & 44.24 & \textcolor{blue}{26.07} & 40.80 & 33.05 & 33.35 & \textbf{\textcolor{red}{25.57}} \\ 
 & CLIPIQA ↑ & 0.5300  & 0.5276 & 0.5225 & 0.5128 & 0.5922 & 0.4993 & \textcolor{blue}{0.6832} & 0.6094 & 0.6729 & 0.6706 & \textbf{\textcolor{red}{0.6941}} \\
 & MUSIQ ↑ & 60.76  & 60.66 & 60.00 & 55.68 & 59.93 & 49.99 & 66.08 & 60.42 & \textbf{\textcolor{red}{67.45}} & 65.73 & \textcolor{blue}{66.85} \\ 
 & BRISQUE ↓ & 17.87  & 22.35 & 24.16 & 21.69 & 19.31 & 23.06 & 15.60 & 17.37 & 23.44 & \textcolor{blue}{15.45} & \textbf{\textcolor{red}{15.28}} \\ 
 & MANIQA ↑ & 0.3552 & 0.3771 & 0.3762 & 0.3192 & 0.3353 & 0.2919 & 0.4231 & 0.4057 & \textcolor{blue}{0.4612} & 0.4570 & \textbf{\textcolor{red}{0.5056} }\\ 
\hline
\multirow{8}{*}{RealSR} & PSNR ↑ & 24.24 & 23.68 & 23.38 & \textbf{\textcolor{red}{24.87} }& 23.13 & \textcolor{blue}{24.35} & 22.82 & 24.27 & 22.93 & 23.05 & 23.30 \\ 
 & SSIM ↑ & \textcolor{blue}{0.7222} & 0.7214 & 0.7167 & \textbf{\textcolor{red}{0.7334}} & 0.6942 & 0.7079 & 0.6635 & 0.6981 & 0.6670 & 0.6201 & 0.6564 \\ 
 & LPIPS ↓ & \textbf{\textcolor{red}{0.2693}} & \textcolor{blue}{0.2753} & 0.2790 & 0.3100 & 0.2928 & 0.2857 & 0.3108 & 0.3494 & 0.3095 & 0.3583 & 0.3323 \\ 
 & FID ↓ & 141.02 & 135.26 & 143.72 & 137.76 & 139.01 & \textcolor{blue}{125.26} & 135.31 & 143.73 & 130.10 & 127.68 & \textbf{\textcolor{red}{123.12}} \\ 
 & CLIPIQA ↑ & 0.5126 & 0.4581 & 0.4561 & 0.3585 & 0.5323 & 0.4817 & 0.6290 & 0.5439 & 0.6274 & \textcolor{blue}{0.6576} & \textbf{\textcolor{red}{0.6582}} \\
 & MUSIQ ↑ & 63.23 & 60.43 & 60.79 & 45.44 & 58.32 & 52.42 & 65.58 & 58.28 & \textbf{\textcolor{red}{67.81}} & 64.14 & \textcolor{blue}{65.83} \\ 
 & BRISQUE ↓ & 29.21 & 31.62 & 32.73 & 43.76 & 27.79 & 36.93 & \textcolor{blue}{18.92} & 27.23 & 24.00 & 20.25 & \textbf{\textcolor{red}{18.33}} \\ 
 & MANIQA ↑ & 0.3761 & 0.3764 & 0.3787 & 0.2726 & 0.3489 & 0.3002 & 0.4325 & 0.3789 & \textcolor{blue}{0.4698} & 0.4433 & \textbf{\textcolor{red}{0.5007}} \\  
\hline
\multirow{8}{*}{DRealSR} & PSNR ↑ & 25.04 & 25.19 & 24.80 & \textbf{\textcolor{red}{26.29}} & 23.51 & \textcolor{blue}{25.87} & 25.23 & 25.19 & 24.58 & 24.86 & 25.66 \\ 
 & SSIM ↑ & 0.7234 & 0.7339 & \textcolor{blue}{0.7396} & \textbf{\textcolor{red}{0.7620}} & 0.6730 & 0.7323 & 0.6944 & 0.6899 & 0.6776 & 0.6184 & 0.6782 \\ 
 & LPIPS ↓ & 0.2960 & \textcolor{blue}{0.2909} & \textbf{\textcolor{red}{0.2885}} & 0.3178 & 0.3261 & 0.3153 & 0.3242 & 0.4083 & 0.3680 & 0.4394 & 0.3875 \\ 
 & FID ↓ & 155.94 & \textbf{\textcolor{red}{145.80}} & 157.44 & 156.45 & 156.10 & \textcolor{blue}{150.08} & 151.35 & 176.03 & 154.17 & 164.23 & 157.51 \\ 
 & CLIPIQA ↑ & 0.5061 & 0.4488 & 0.4470 & 0.3796 & 0.5620 & 0.4622 & 0.6224 & 0.5272 & 0.6424 & \textcolor{blue}{0.6559} & \textbf{\textcolor{red}{0.6616} }\\
 & MUSIQ ↑ & 57.12 & 54.28 & 54.01 & 42.48 & 53.64 & 41.36 & 58.28 & 49.73 & \textbf{\textcolor{red}{63.25}} & 60.27 & \textcolor{blue}{60.38}\\ 
 & BRISQUE ↓ & 32.69 & 34.90 & 39.06 & 43.31 & 26.36 & 38.94 & 20.42 & 25.23 & 25.50 & \textbf{\textcolor{red}{17.55}} & \textcolor{blue}{18.62} \\ 
 & MANIQA ↑ & 0.3419 & 0.3426 & 0.3442 & 0.2840 & 0.3151 & 0.2701 & 0.3801 & 0.3255 & \textcolor{blue}{0.4568} & 0.4530 & \textbf{\textcolor{red}{0.4608} }\\  
 \hline
\end{tabular}%
}
\label{tab:Quantitative Comparisons}
\end{table}

\begin{figure}[tb]
\vspace{-2mm}
  \centering
  \includegraphics[height=5.5cm]{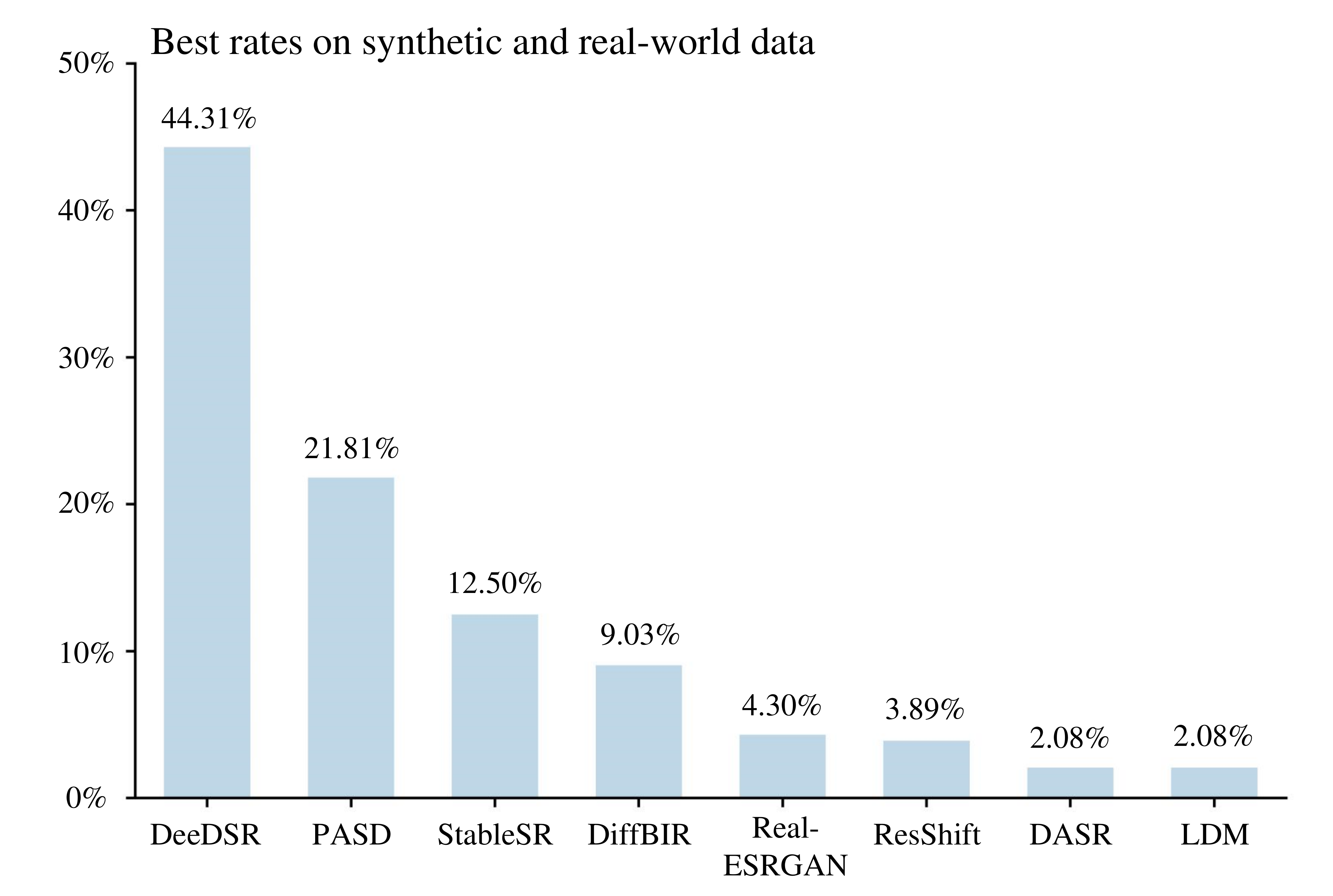}
  \caption{Results of user study on synthetic and real-world data.}
  \label{fig:user study}
\end{figure}

\begin{figure}[h]
  \centering
  \includegraphics[height=5cm]{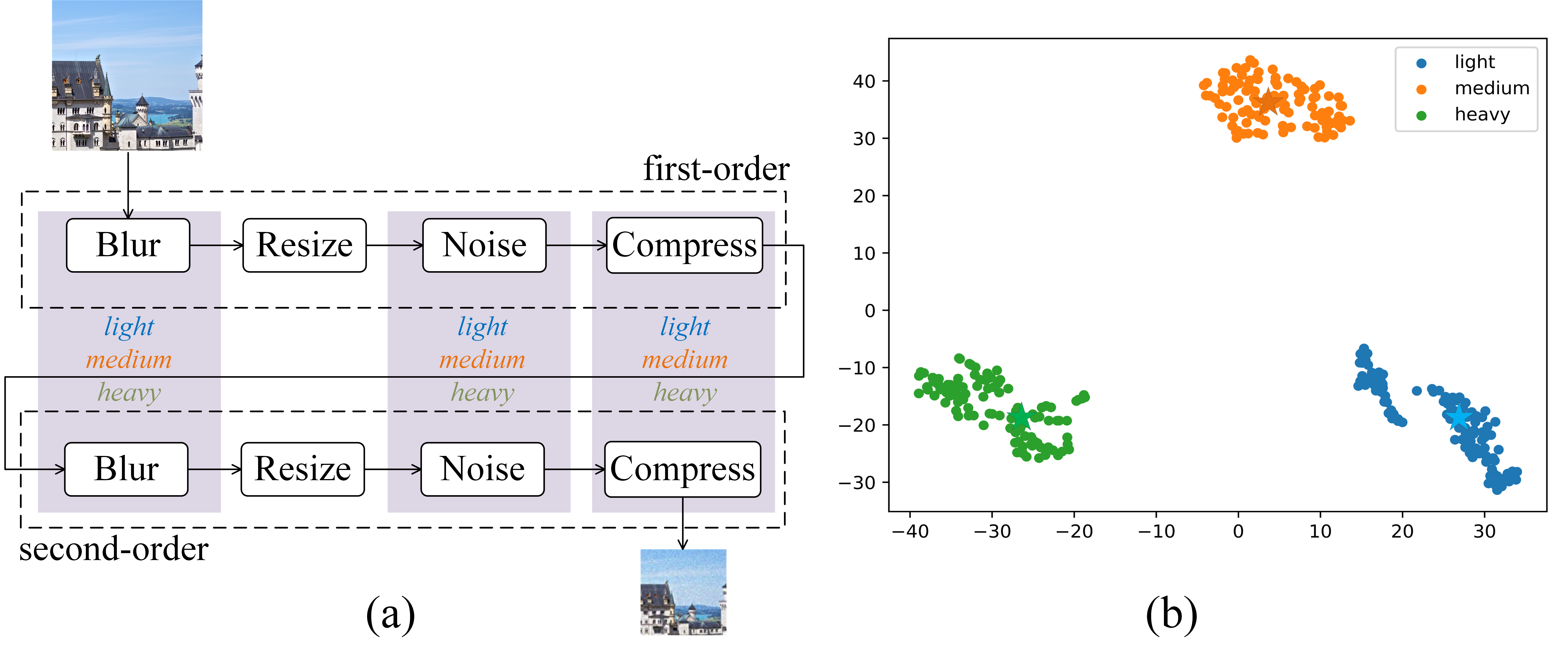}
  \caption{(a) Description of the method for distinguishing degradation levels as light, medium, and heavy. (b) Visualization of degradation representations estimated by the Degradation Learner. Please refer to more details in the \textbf{supplementary material}. 
  }
  \vspace{-2mm}
  \label{fig:degradation}
\end{figure}

\vspace{-2mm}
\subsection{Comparisons with State-of-the-Arts}
\noindent\textbf{Quantitative Comparisons.}
We conduct quantitative comparisons across three synthetic and real datasets, as illustrated in \cref{tab:Quantitative Comparisons}. 
Our method achieves superior scores on all datasets across various perceptual metrics, including CLIPIQA and MANIQA. We rank second to PASD on the MUSIQ metric. Additionally, our BRISQUE score is highest on both DIV2K-Val and RealSR datasets and is only 6\% lower than PASD on DRealSR.

Notably, GAN-based methods outperform diffusion-based methods in reference metrics due to diffusion models' ability to generate richer texture details, albeit with a slight reduction in fidelity. Our method achieves the highest FID score on both DIV2K-Val and RealSR, along with less than 0.1\% LPIPS score on DIV2K-Val after StableSR. In contrast to LDM and ResShift, our DeeDSR consistently achieves the highest PSNR score among ControlNet-based methods like StableSR, PASD, and DiffBIR. Overall, we achieve better image generation quality and competitive fidelity compared to diffusion-based methods.

\noindent\textbf{Qualitative Comparisons.}
\cref{fig:Qualitative Comparisons(synthetic)} and \cref{fig:Qualitative Comparisons(real)} provide visual comparisons of various methods on synthetic (DIV2K-Val) and real-world datasets (RealSR, DRealSR), respectively. DeeDSR demonstrates superior image generation quality and fidelity across the examples. On the synthetic dataset, DeeDSR accurately restores details like tentacles, leaf textures, faces without distortion, and architectural structures, while other methods yield blurry, noisy, or semantically incorrect outputs. On real-world datasets, DeeDSR uniquely produces accurate glyphs, stripes, and jagged textures resembling the ground truth, showcasing its ability to provide precise and comprehensive semantics for SR.



\noindent\textbf{User Study.}
To further demonstrate the effectiveness of our DeeDSR, we conduct a user study on both synthetic and real-world datasets. We randomly choose 10 samples from DIV2K-Val, 5 samples from RealSR, and 5 samples from DRealSR. 36 participants evaluated seven representative methods and our method: Real-ESRGAN, DASR, LDM, StableSR, ResShift, PASD, DiffBIR, and Our DeeDSR. Giving the LR image and the corresponding HR image as reference, the participants were asked to answer ‘Which image is the best SR result of the LR image among the eight restored results?’ We obtained $20\times36$ votes in total and calculated the selection rate as depicted in \cref{fig:user study}. Our DeeDSR achieved a selection rate of 44.31\%, approximately twice as high as the second-ranked method.

\noindent\textbf{Semantics Preservation Test}
To evaluate the semantic fidelity capabilities of various methods, we adopt OpenSeed \cite{Zhang_2023_openseed} to evaluate the SR images generated by different methods for image detection and segmentation tasks. The COCO dataset validation set (COCO-Val) consisting of 5,000 images is used for evaluation. We resize the COCO-Val images to 512$\times$512  and follow \cite{wang2021realesrgan} to generate the corresponding 
LR images of  $128\times128$. We then produce the SR images using different methods. As shown in \cref{tab:semantic_restoration_comparison}, our method achieves the highest scores across four representative metrics, demonstrating superior semantic fidelity.

\begin{table}[t]
\centering
\caption{The comparisons of semantic restoration performance among different SR methods.}
\resizebox{\textwidth}{!}{%
\begin{tabular}{@{}lcccccccccccccc@{}}
\toprule
\multirow{2}{*}{Metrics} & \multicolumn{13}{c}{Methods} \\ \cmidrule(lr){2-14} 
                          & Zoomed LR & BSRGAN & Real-ESRGAN & LDL & DASR & FeMaSR & LDM & StableSR & ResShift & PASD & DiffBIR & DeeDSR & GT \\ \midrule
Panoptic Segmentation (PQ) & 24.3 & 23.5 & 27.1 & 25.5 & 22.5 & 23.3 & 22.6 & 33.8 & 28.6 & 33.4 & 27.2 & \textbf{\textcolor{red}{35.8}} & 55.4 \\
Object Detection (AP)     & 17.5 & 16.2 & 19.6 & 18.2 & 15.7 & 16.2 & 15.4 & 24.9 & 21.1 & 23.9 & 19.3 & \textbf{\textcolor{red}{26.9}} & 52.3 \\
Instance Segmentation (AP) & 15.4 & 14.3 & 17.4 & 16.1 & 13.8 & 14.2 & 13.4 & 21.7 & 18.5 & 20.8 & 17.1 & \textbf{\textcolor{red}{23.4}} & 47.1 \\
Semantic Segmentation (mIOU) & 32.2 & 30.7 & 36.7 & 34.1 & 29.8 & 29.3 & 29.8 & 45.4 & 38.6 & \textbf{\textcolor{red}{47.9}} & 33.8 & \textbf{\textcolor{red}{47.9}} & 63.8 \\ \bottomrule
\end{tabular}%
}
\vspace{-2mm}
\label{tab:semantic_restoration_comparison}
\end{table}

\begin{table}[t]
\centering
\renewcommand\arraystretch{1.2} 
\renewcommand{\tabcolsep}{3pt} 
\caption{Ablation studies on DIV2K-Val and RealSR benchmarks for the Real-ISR task. For variant (4), we generate 3000 HR-LR pairs using the same configuration with text prompts for evaluation (denoted as DIV2K-Val-V2). This evaluation is not conducted on RealSR due to the absence of text prompts.}
\label{tab:ablation_studies}
\resizebox{0.98\textwidth}{!}{%
\begin{tabular}{@{}c|cc|cc|c|cccc@{}}
\hline
\multirow{2}{*}{Variants} & \multicolumn{2}{c|}{Prompt Type}       & \multicolumn{2}{c|}{Control Form}   & \multirow{2}{*}{\begin{tabular}[c]{@{}c@{}}Degradation\\ Learner\end{tabular}} & \multicolumn{4}{c}{DIV2K-Val/RealSR} \\ \cline{2-5} \cline{7-10} 
                  & \multicolumn{1}{c|}{Image} & Text & \multicolumn{1}{c|}{Global} & Local & & \multicolumn{1}{c}{PSNR $\uparrow$ }   & \multicolumn{1}{c}{LPIPS $\downarrow$  } & \multicolumn{1}{c}{FID $\downarrow$}  & \multicolumn{1}{c}{MUSIQ $\uparrow$}  \\ \hline
(1) & \checkmark  & & \checkmark  & \checkmark  &  &  22.00/22.71  &   0.3423/0.3423 &   31.27/141/33 &  \textbf{\textcolor{red}{66.90}}/\textbf{\textcolor{red}{68.90}}       \\
(2)  & \checkmark & &  & \checkmark & \checkmark  &          22.25/23.02  &  0.3337/0.3366  & 30.30/134.66  &  66.35/68.24    \\
(3)  & \checkmark  & & \checkmark &   & \checkmark   &      17.66/12.79   & 0.5368/0.7957 &   87.84/275.08   &   65.58/64.70      \\
DeeDSR  & \checkmark &  & \checkmark  & \checkmark& \checkmark &   \textbf{\textcolor{red}{22.51}}/\textbf{\textcolor{red}{23.30} } &          \textbf{\textcolor{red}{0.3133}}/ \textbf{\textcolor{red}{0.3323}} &   \textbf{\textcolor{red}{25.57}}/\textbf{\textcolor{red}{123.12}}  &   66.85/65.83  \\ \hline
(4)  &  & \checkmark   & \checkmark & \checkmark    &  &  22.50/-       &   0.3284/-       &  28.54/-     &   66.03/-      \\
DeeDSR  &\checkmark &  & \checkmark & \checkmark     & \checkmark  &   \textbf{\textcolor{red}{22.56}}/- &    \textbf{\textcolor{red}{0.3123}}/- &    \textbf{\textcolor{red}{26.11}}/-  &   \textbf{\textcolor{red}{66.86}}/-      \\ \hline

\end{tabular}}
\end{table}

\subsection{Ablation Study}
\label{sec:ablation}
\noindent\textbf{Effectiveness of the Degradation Learner.}
To evaluate the efficacy of our degradation representation compared to text-based degradation, we adopt the network structure from \cite{chen2023textprompt}, excluding the first stage and DA blocks, referred to as variant (4). Instead of using our degradation representation, we describe the degradation process through text (e.g., blur, resize, noise, JPEG compression) and feed it into the Stable Diffusion following \cite{chen2023textprompt}. We also create a synthetic dataset with corresponding degradation texts on the DIV2K validation set (DIV2K-Val-V2) for evaluation. The results of variant (4) presented in \cref{tab:ablation_studies} indicate that image prompts maintain more accurate degradation representations than text prompts, enabling the model to discern correct image semantics.

To evaluate the effectiveness of Degradation Learner across varying degradation levels, we classify LR images into three groups based on the level of degradation, \textit{i.e.}, light, medium, and heavy as shown in \cref{fig:degradation}(a). For specific details, please refer to the \textbf{supplementary material}.
Using a pre-trained Degradation Learner, we obtain degradation representations for these groups and visualize them with T-SNE \cite{van2008tsne} in \cref{fig:degradation}(b). The Degradation Learner generates discriminative clusters, distinguishing different degradation levels.

\cref{fig:introduction} compares DeeDSR and StableSR across degradation levels. StableSR suffers from blurring and incorrect textures under heavy degradation, which DeeDSR effectively addresses due to its degradation-aware capabilities.

To assess the influence of the Degradation Learner, we exclude it and the DA blocks of DeeDSR, which corresponds to variant (1). As shown in \cref{tab:ablation_studies}, DeeDSR outperforms variant (1) by approximately 8\% on both synthetic and real-world datasets in terms of the reference metrics. While the MUSIQ perceptual metric difference is negligible on DIV2K-Val, DeeDSR scores approximately 4\% lower than variant (1) on RealSR. This underscores the Degradation Learner with DA blocks' effectiveness in aiding the Degradation-Aware Adapter to extract precise image semantics, leading to improved fidelity but a slight reduction in perceptual metric scores, attributed to the suppression of some unnecessary texture details.

\noindent\textbf{Effectiveness of the Global Representation.} The impact of removing the global representation from DeeDSR, denoted as variant (2), is shown in \cref{tab:ablation_studies}. Compared to variant (4), incorporating global semantic information from the low-resolution input aids Stable Diffusion's understanding of the real-world super-resolution task, thereby improving overall image fidelity.

\noindent\textbf{Effectiveness of the Local Representation.} Based on our framework, the local representation acts as the primary content guide to control the generation of SD, while the global representation involves integrating the high-dimensional semantic of LR into the cross-attention module. This module promotes global semantic understanding to guide generation, but lacks providing content information. Therefore, removing the local representation, i.e., variant (3), results in a significant decrease in model performance, as shown in \cref{tab:ablation_studies}..


\subsection{Complexity Comparison}
\begin{table}[h]
\centering
\caption{Complexity comparison of model parameters and running time. All methods are evaluated on 128 × 128 input images for 4x SR using an NVIDIA RTX 3090 GPU.}
\renewcommand\arraystretch{1.1}
\resizebox{0.94\textwidth}{!}{%
\begin{tabular}{@{}ccccccccc@{}}
\toprule
 &  \textbf{Real-ESRGAN}\cite{wang2021realesrgan}  & \textbf{FeMaSR}\cite{chen2022real} & \textbf{LDM}\cite{rombach2022high}   & \textbf{StableSR}\cite{wang2023stablesr}   & \textbf{ResShift} \cite{yue2024resshift}  & \textbf{PASD} \cite{yang2023pasd}
& \textbf{DiffBIR}\cite{lin2023diffbir} & \textbf{DeeDSR} 
\\ \hline
 Model type  & GAN  & GAN & LDM & LDM & LDM & LDM & LDM & LDM \\
 Time steps  & 1  & 1 & 50 & 200 & 15 & 20 & 50 & 50 \\
 Runtime& 0.04s  & 0.05s & 1.85s & 10.90s & 0.71s & 4.07s &  5.35s & 7.13s \\
 Params  & 16.7M   & 28.3M & 169.0M & 1409.1M & 173.9M & 1900.4M & 1716.7M &  1017.6M\\
\bottomrule
\end{tabular}%
}
\label{tab:table4}
\end{table}

We conduct a comparative analysis of various methods on the DIV2K-Val dataset, evaluating inference time and parameter counts. The results are illustrated in \cref{tab:table4}. The inference time is measured for reconstructing a 512 × 512 HR image from 128 × 128 input using an NVIDIA RTX 3090 GPU. GAN-based SR methods exhibit shorter inference times compared to diffusion-based methods due to reduced forward pass and fewer parameters. Among diffusion-based methods, LDM and ResShift demonstrate significantly faster inference speed owing to minimized parameterization. Despite higher parameter counts, PASD and DiffBIR exhibit relatively faster inference speed than StableSR, which requires more sampling steps. Our proposed DeeDSR maintains fewer parameters but is slightly slower than PASD and DiffBIR due to an extra stage for estimating degradations. Overall, DeeDSR outperforms state-of-the-art approaches in visual quality and quantitative results with reasonable inference time and parameters.

\vspace{-2mm}
\section{Conclusion}
\vspace{-2mm}
This paper proposes DeeDSR, a novel approach that leverages degradation-aware prompts to enhance pre-trained T2I models for real-world image super-resolution. DeeDSR captures global degradation representations while ensuring semantic accuracy through a two-stage process: contrastive learning to grasp degradation nuances, and integrating these insights with low-resolution images to precisely guide the T2I model. This strategy generates highly detailed and semantically coherent images. Extensive experiments demonstrate DeeDSR's superiority in image recovery across various degradations, outperforming existing methods.


%
%




{\small
\bibliographystyle{splncs04}
\bibliography{main}
}

\clearpage
\appendix


This supplementary material provides additional details not included in the main paper due to space constraints. \cref{sec:arch_adapter} presents the details of our proposed Degradation-Aware Adapter. \cref{sec:ablation_descrip} provides further descriptions for our ablation studies. \cref{sec:addition_results} showcases more visual results on synthetic and real-world datasets. \textbf{We also include a local web page for SR results from our DeeDSR. Readers are encouraged to view this web page (open index.html using Microsoft Edge browser) for better observation and comparison.}

\section{Architecture of Degradation-Aware Adapter}
\label{sec:arch_adapter}
As described in the main paper, we reduce the number of modules in the original ControlNet ~\cite{zhang2023adding} to reduce computational complexity, decreasing the number of trainable parameters by approximately one-third. The Degradation-Aware Adapter has 75.90 M parameters. The detailed settings are listed in \cref{tab:Settings}.
To incorporate the LR features and the estimated degradation representations, we introduce the DA block between the residual block and the cross-attention layer in each scale of the Degradation-Aware Adapter. 
The detailed architecture of DA block is illustrated in \cref{{fig:DA block}}.

\begin{figure}[h]
\vspace{-2mm}
  \centering
  \includegraphics[height=5.5cm]{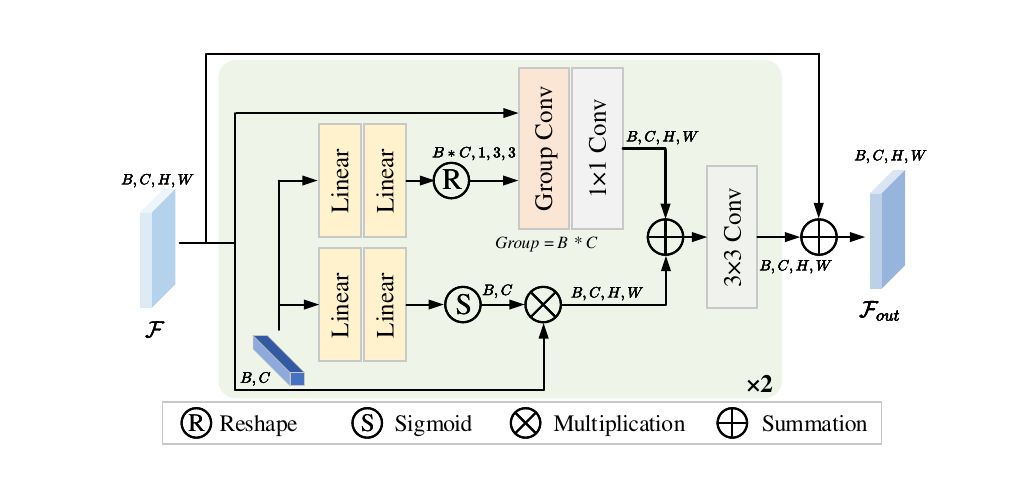}
  \caption{Detailed architecture of the DA block. }
  \label{fig:DA block}
\end{figure}

\begin{table}[h!]
\vspace{-4mm}
\centering
\caption{Settings of the Degradation-Aware Adapter in DeeDSR.}
\renewcommand{\tabcolsep}{3.5pt} 
    \renewcommand\arraystretch{1.0}
\begin{tabular}{lcc}
\toprule
\textbf{Stage} & \textbf{Block} & \textbf{Output Size} \\
\midrule
Input Block 1 & Conv(kernel\_size=3) & \( H \times W \times 256 \)  \\
\hline
 \multirow{4}{*}{Input Block 2} & ResBlock & \\
 & DA Block & \\
 & AttentionBlock & \\
 & ResBlock(down=True) & \multirow{-4}{*}{\( H/2 \times W/2 \times 256 \)} \\
\hline
 \multirow{4}{*}{Input Block 3} & ResBlock & \\
 & DA Block & \\
 & AttentionBlock & \\
 & ResBlock(down=True) & \multirow{-4}{*}{\( H/4 \times W/4 \times 256 \)} \\
\hline
 \multirow{4}{*}{Input Block 4} & ResBlock & \\
 & DA Block & \\
 & AttentionBlock & \\
 & ResBlock(down=True) & \multirow{-4}{*}{\( H/8 \times W/8 \times 512 \)} \\
\hline
\multirow{3}{*}{Input Block 5} & ResBlock & \\
 & DA Block & \\
 & AttentionBlock & \multirow{-3}{*}{\( H/8 \times W/8 \times 512 \)} \\
  
\hline
\multirow{3}{*}{Middle Block} & ResBlock & \\
 & AttentionBlock & \\
 & ResBlock & \multirow{-3}{*}{\( H/8 \times W/8 \times 512 \)} \\
\bottomrule
\end{tabular}
\label{tab:Settings}
\end{table}


\begin{table}[h]
\centering
    \caption{Synthetic of Different Degradation Levels. $\sigma$: Blur Sigma; ${\beta_g}$: Betag Range; ${\beta_p}$: Betap Range; $\varepsilon$: Noise Level; ${q}$: JPEG Quality Range. }
    \renewcommand{\tabcolsep}{3.5pt} 
    \renewcommand\arraystretch{1.0}

\begin{tabular}{c|c|c|cc|cc|cc}
\hline
\multicolumn{1}{c|}{\multirow{2}{*}{}} & \multicolumn{1}{c|}{\multirow{2}{*}{Degradation}} & \multirow{2}{*}{Param.} & \multicolumn{2}{c|}{Light} & \multicolumn{2}{c|}{Medium}    & \multicolumn{2}{c}{Heavy}                         \\ \cline{4-9} 
\multicolumn{1}{c|}{}  & \multicolumn{1}{c|}{}   &  & min  & max         & min & \multicolumn{1}{c|}{max} & \multicolumn{1}{c}{min} & \multicolumn{1}{c}{max} \\ \hline
\multirow{5}{*}{\begin{tabular}[c]{@{}c@{}}First\\ order\end{tabular}}  & \multirow{3}{*}{Blur} & ${\sigma _1}$  &    0.2  &  0.6      &  0.6 & 1.1&1.1 & 1.5  \\
   &    & ${\beta _{g1}}$ &   0.5  & 1.0   & 1.0 &1.5 &1.5 &  2.0  \\
  &     & ${\beta _{p1}}$ &   1.0    &   1.2     &  1.2   & 1.4&1.4 &1.5    \\ \cline{2-9} 
     & Noise      & ${\varepsilon _1}$  & 1.0      &  5.0      &  5.0 & 10.0&10.0&  15.0    \\ \cline{2-9} 
      &Compression  & ${q_1}$  &   81    &     95  &  70  &81& 60&  70   \\ \hline
\multirow{2}{*}{\begin{tabular}[c]{@{}c@{}}Second\\ order\end{tabular}}  & Noise    & ${\varepsilon _2}$  &   1.0    &  4.0      &  4.0   &8.0 &8.0 & 12.0   \\ \cline{2-9}  
      & Compression   & ${q_2}$  &   86    &      100  &  73  &86 & 60&  73    \\ \hline
\end{tabular}

\label{tab:Synthetic}
\end{table}

\section{Additional Description for Ablation Study}
\label{sec:ablation_descrip}
\noindent\textbf{Synthetic of Different Degradation Levels}
To evaluate the effectiveness of Degradation Learner across varying degradation levels, we classify LR images into three groups based on the level of degradation, \textit{i.e.}, light, medium, and heavy. Specifically, we maintain the non-parameterized degradation process constant throughout, by eliminating the resizing operation, setting the probability of blur in the second stage to 0, and fixing the probability of Gaussian noise to 1. Subsequently, the other parameterizable operations are evenly distributed into light, medium, and heavy categories within the parameter interval as shown in \cref{tab:Synthetic}.

\begin{figure}[h]
\vspace{-2mm}
  \centering
  \includegraphics[width=7cm]{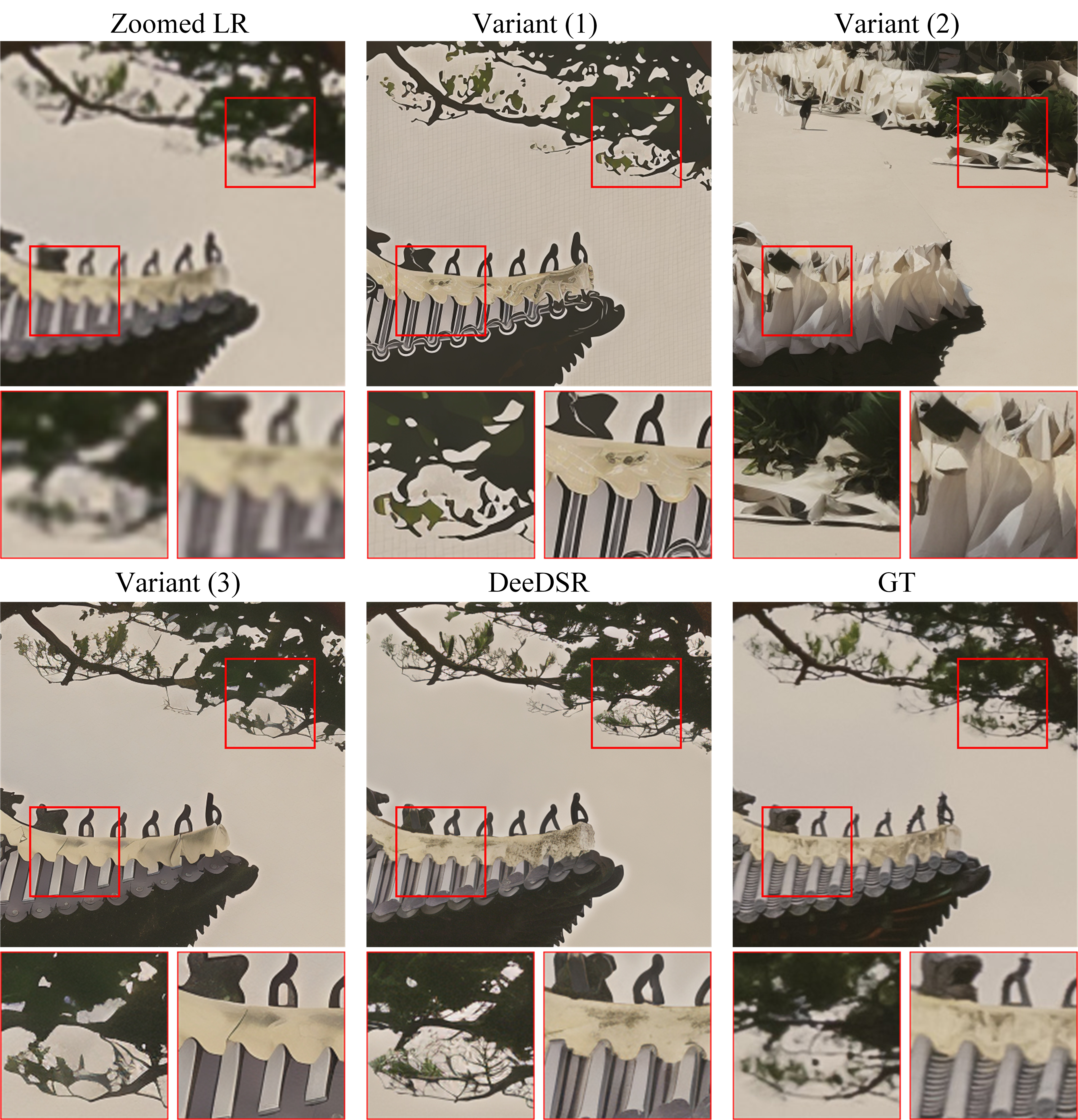}
  \caption{Ablation results of variants (1), (2), (3) on the real-world dataset.  }
  \label{fig:ablation1}
\end{figure}

\begin{figure}[h]
\vspace{-2mm}
  \centering
  \includegraphics[width=7cm]{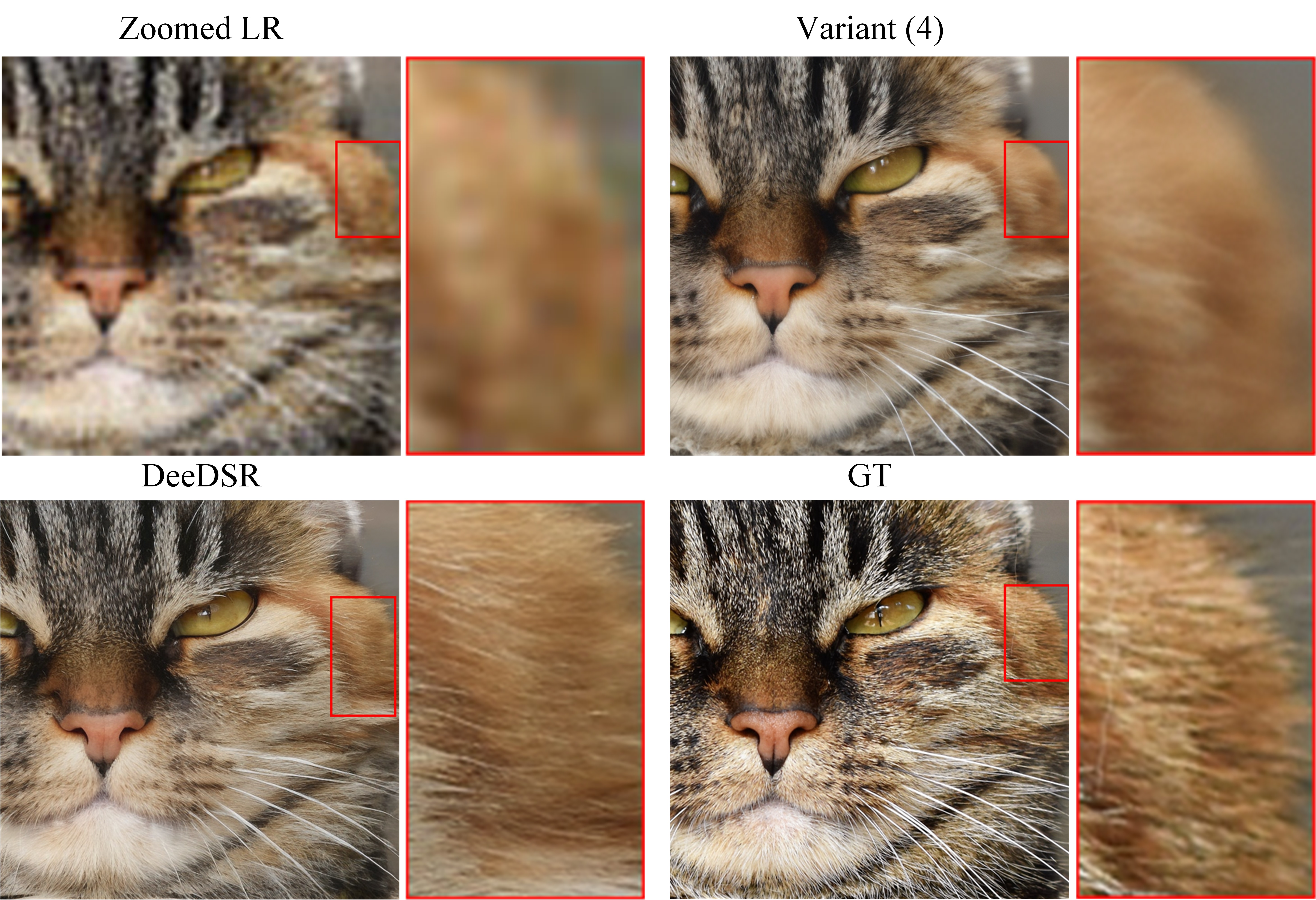}
  \caption{Ablation result of variant (4) on the synthetic dataset.  }
  \label{fig:ablation2}
\end{figure}
\noindent\textbf{Qualitative Comparison of Variants.}
We present visual results of different variants in \cref{fig:ablation1} and \cref{fig:ablation2}, where variants (1), (2), (3) and (4) correspond to the configurations outlined in the main paper.

As illustrated in \cref{fig:ablation1}, variant (1) removes the Degradation Learner module, which fails to perceive degradation, leading to incorrect or incomplete local and global semantics in the generated outputs. It mainly manifests as mistextured images. Variant (2) removes the local branch, causing a lack of local perception and fidelity in the generated results. In Variant (3), the removal of the global semantic control branch from DeeDSR results in a lack of certain detail information that would typically be provided by semantic guidance. For instance, the generated images fail to recover intricate details such as tree branches and roof tiles present in the original input. The absence of this branch leads to a loss of fidelity in capturing the nuanced local and global semantics inherent in the data. Variant (4) replaces the image with a text prompt with the appropriate structure, resulting in a low-quality generated image with inaccurate degradation information, as shown in ~\cref{fig:ablation2}.

The ablation results indicate that our DeeDSR incorporates high-dimensional semantics provided by the image prompt along with local details, effectively enhancing SD's generative capacity and realism of outputs. Additionally, it maintains the overall style consistency with the LR image, thereby improving fidelity in terms of luminance, contrast, and color features.

\noindent\textbf{Noise Guidance Strategies.}
We conduct ablation studies on DeeDSR with three distinct inference strategies: Noise Guidance (default configuration defined in the main paper), LR Initialization (Embedding LR into Initial Noise), implemented using the following equation:
\begin{equation}
z_t = \sqrt{\overline{\alpha}_t} z + \sqrt{1 - \overline{\alpha}_t} \varepsilon,
\end{equation}
and the original strategy that starts sampling from random Gaussian noise (Noise Initialization). The results in  \cref{tab:ablation_noiseguidance} reveal that LR Initialization yields higher fidelity, as evidenced by its superior performance on reconstruction metrics like PSNR, but it sacrifices realism and perceptual quality, as indicated by its lower scores on generative metrics such as FID and MUSIQ. Conversely, Noise Initialization demonstrates the opposite trend, achieving better realism and perceptual quality at the cost of fidelity, as reflected in its higher FID and MUSIQ scores but lower PSNR values. Our proposed Noise Guidance method strikes a better balance between fidelity and realism, outperforming LR Initialization on perceptual metrics like LPIPS and MUSIQ while maintaining competitive performance on reconstruction metrics like PSNR, indicating its ability to generate realistic and visually appealing images.


\begin{table}[h]
\centering
\renewcommand\arraystretch{1.0} 
\renewcommand{\tabcolsep}{3pt} 
\caption{Ablation Studies of Different Inference Strategies.}
\label{tab:ablation_noiseguidance}

\resizebox{0.8\textwidth}{!}{%
\begin{tabular}{@{}c|cccc@{}}
\hline
\multirow{2}{*}{Strategy}  & \multicolumn{4}{c}{DIV2K-Val/RealSR} \\ \cline{2-5} 
                  & \multicolumn{1}{c}{PSNR $\uparrow$ }   & \multicolumn{1}{c}{LPIPS $\downarrow$  } & \multicolumn{1}{c}{FID $\downarrow$}  & \multicolumn{1}{c}{MUSIQ $\uparrow$}  \\ \hline
Noise Guidance &  22.51/23.60  &   \textbf{\textcolor{red}{0.3133}}/0.3198 &   25.57/125.62 &  66.85/64.48       \\
LR Initialization  & \textbf{\textcolor{red}{23.16}}/\textbf{\textcolor{red}{24.17} } &  0.3350/\textbf{\textcolor{red}{0.2857}}  & 27.82/124.22  &  60.31/61.54    \\
Noise Initialization  & 22.33/23.30   & 0.3176/0.3323 &   \textbf{\textcolor{red}{25.33}}/\textbf{\textcolor{red}{123.12}}   &   \textbf{\textcolor{red}{67.34}}/\textbf{\textcolor{red}{65.83}}      \\
\hline

\end{tabular}}

\end{table}

\begin{figure}[h]
\vspace{-2mm}
  \centering
  \includegraphics[width=1\textwidth]{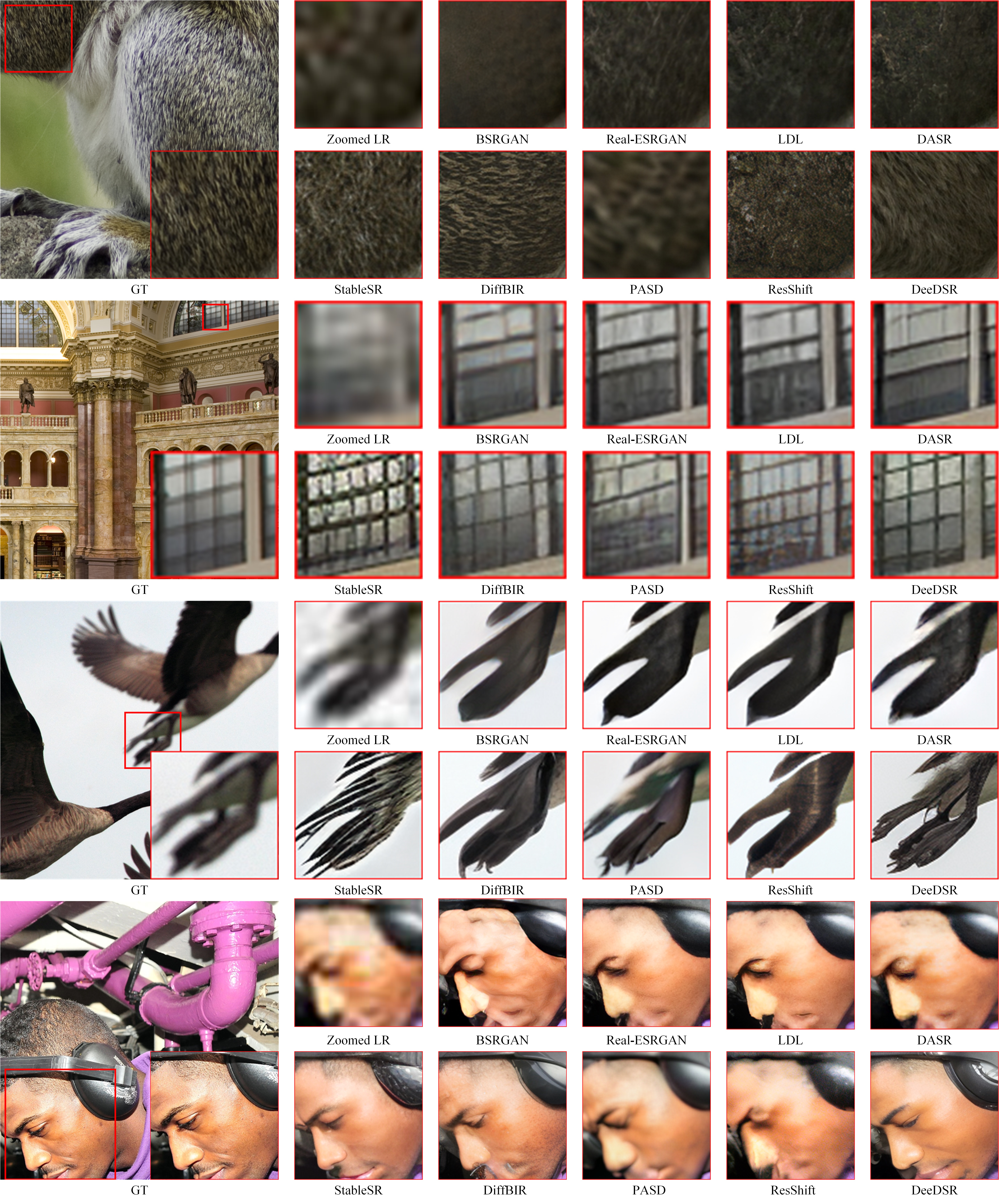}
  \caption{Qualitative comparison with different methods on synthetic examples. Our method can accurately restore the texture and details of the corresponding object under challenging degradation. Zoom in for a better view.}
  \label{fig:ablation3}
\end{figure}

\begin{figure}[h]
\vspace{-2mm}
  \centering
  \includegraphics[width=\textwidth]{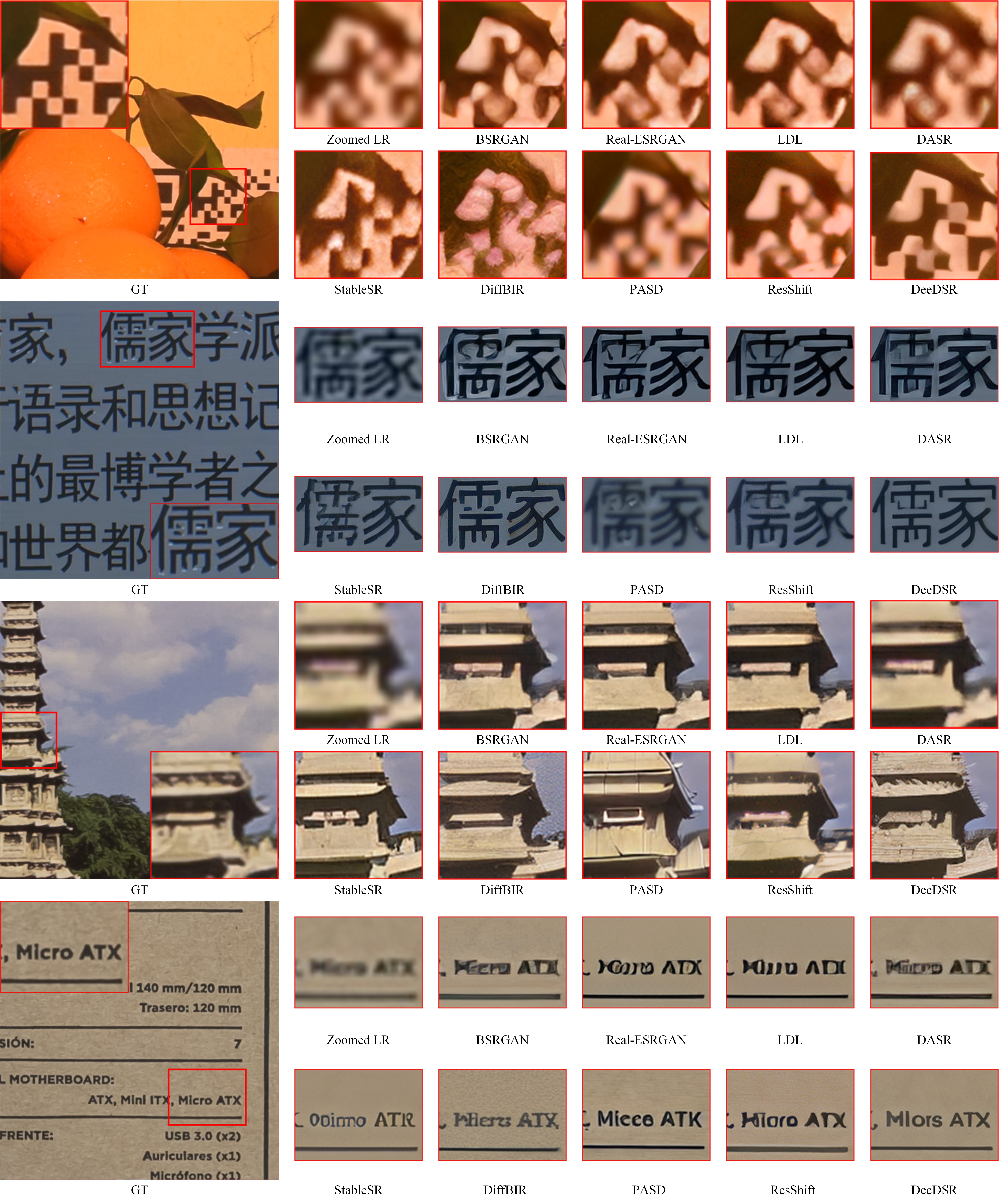}
  \caption{Qualitative comparisons of different methods on real-world datasets. Please zoom in for a better view.  }
  \label{fig:ablation4}
 
\end{figure}

\section{Additional Comparison Results}
\label{sec:addition_results}
To verify the effectiveness,  we provide more qualitative results in \cref{fig:ablation3} and \cref{fig:ablation4} compared to several SOTA GAN-based and Diffusion-based SR methods, \textit{i.e.}, BSRGAN\cite{zhang2021designing}, Real-ESRGAN\cite{wang2021realesrgan}, LDL\cite{liang2022details}, DASR\cite{liang2022efficient}, StableSR \cite{wang2023stablesr}, DiffBIR~\cite{lin2023diffbir}, PASD~\cite{yang2023pasd}, ResShift\cite{yue2024resshift}. Numerous instances demonstrate the powerful restoration capability of our proposed DeeDSR, generating highly realistic restored images.

In addition, we use the same aggregation sampling strategy with \cite{wang2023stablesr} to sample images of arbitrary resolutions, and the results are presented in our local
web page.


\end{document}